\documentclass{article}
\usepackage{comment}
\usepackage[paperheight=11. 5in, paperwidth=9. 5in]{geometry}
\usepackage{lscape}
\usepackage{ amssymb }
\usepackage[utf8]{inputenc}
\usepackage[final]{pdfpages}
\usepackage{amsmath}
\usepackage{slashbox }
\usepackage[libertine]{newtxmath}
\usepackage{color}
\usepackage{setspace}
\usepackage{pdflscape}
\usepackage{graphicx} 
\usepackage{float}
\DeclareUnicodeCharacter{FB01}{fi}
\usepackage{algorithm}
\usepackage{algorithmicx}
\usepackage{algpseudocode}

\title{An Introduction to Advanced Machine Learning :   Meta Learning Algorithms,  Applications and  Promises }

\author{Farid Ghareh Mohammadi$^1$,  M.  Hadi  Amini$^2$,  and Hamid R.  Arabnia$^1$\\1 :  Department of Computer Science,  Franklin College of Arts and Sciences, \\ University of Georgia,  Athens,  Georgia,  30601  \\
2 :  School of Computing and Information Sciences,  College of Engineering and Computing,  \\Florida International University,  Miami,  FL 33199 \\
Emails :  farid.ghm@uga.edu,    amini@cs.fiu.edu,  hra@cs.uga.edu}
\date{}

\begin{document}

\maketitle

\section*{Abstract}
 In  \cite{ch1_farid,  ch2_farid},  we have explored the theoretical aspects of feature extraction optimization processes for solving large-scale problems and overcoming machine learning limitations.  Majority of optimization algorithms that have been introduced in  \cite{ch1_farid,  ch2_farid} guarantee  the optimal performance of supervised learning,  given  offline and discrete data,  to deal with curse of dimensionality (CoD) problem.  These algorithms,  however,  are not tailored for solving emerging learning problems.  One of the important issues caused by online data is  lack of sufficient samples per class.  Further,  traditional machine learning algorithms  cannot achieve accurate training based on limited distributed data,  as data has proliferated and dispersed significantly.  Machine learning employs a strict model or embedded engine to train and predict which still fails to learn unseen classes and sufficiently use online data.  In this chapter,  we introduce these challenges elaborately.  We further investigate Meta-Learning (MTL) algorithm,  and their application and promises to solve the emerging problems by answering \textit{how autonomous agents can learn to learn?}.

\textbf{Keywords : } Meta learning,  machine learning,  online learning,  online optimization,   model-based learning,  metric-based learning,  gradient descent,  low shot learning,  few shot learning,  one shot learning

\section{Introduction}

 
 Machine learning algorithms enable researchers to learn from supervised / unsupervised data.  Collected data is mainly offline  and it is not evolving over time.  Hence,   total behavior of future data are vague enough for us to process.  Conventionally,   it is not possible to have the entire behaviour learned  \cite{Schmidhuber1987Learning} using the traditional machine learning,  evolutionary algorithms and optimization algorithms discussed earlier  \cite{ch1_farid, ch2_farid} . 
\begin{figure}[H]
    \centering
    \includegraphics[height=3.5in]{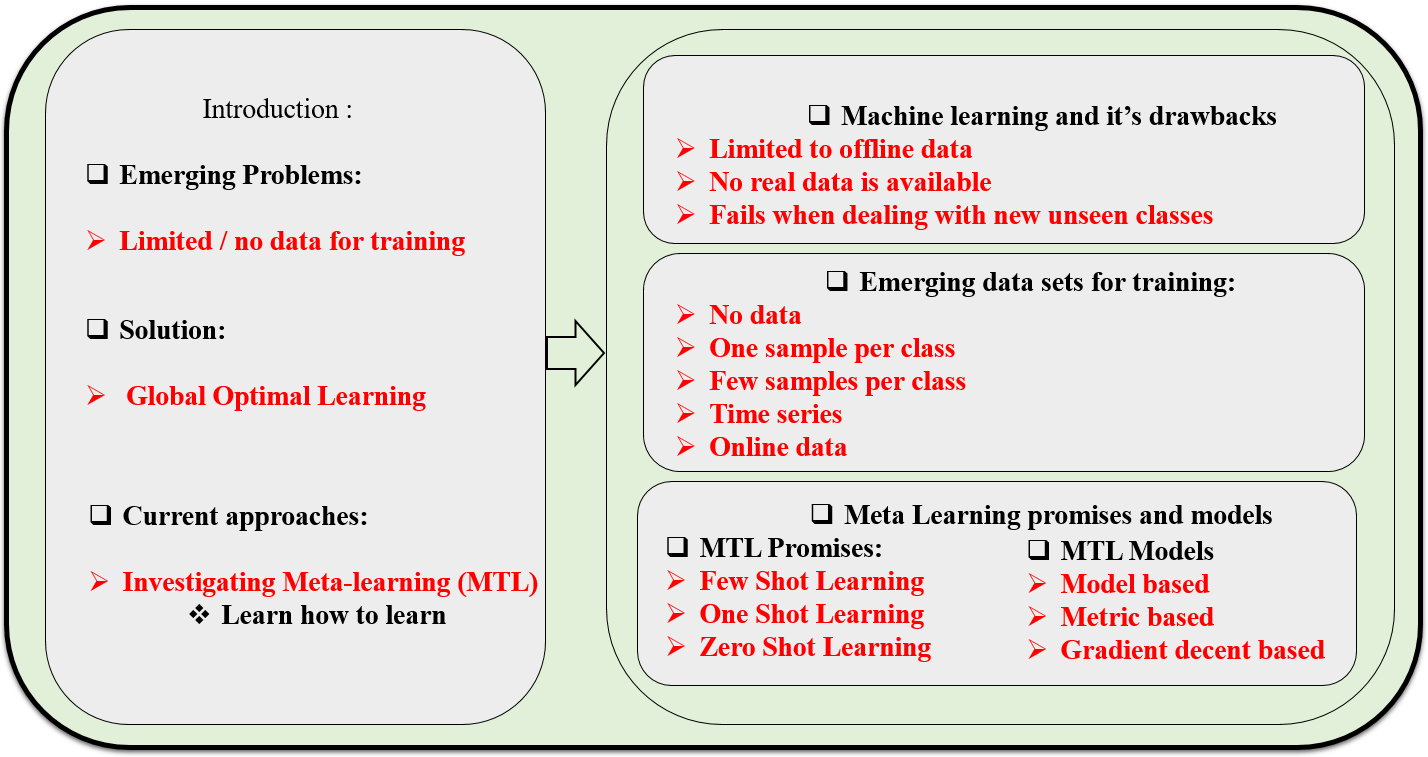}
    \caption{Overall structure of this study}
    \label{fig : General_Structure}
\end{figure}

Last decade,  researchers have studied advanced research paradigms to solve learning process.  They aim to to learn using prior tasks or experiences and leverage them for future learning.  One of the promising paradigm is Meta-learning (MTL).  Prior studies investigated MTL methods that learn to update a function or learning rule   \cite{Schmidhuber1987Learning,  Ravi2017fewshot}.  MTL differs from classic machine learning with respect to the level of adaptation \cite{melsurvey01}.   MTL is the process of learning to learn.  It leverages past experiences to ascertain a prior model's parameters and learning process i. e.  algorithm.  MTL investigates how to choose the right bias non-fixed,  unlike base-learning where the bias is fixed a priori  \cite{melsurvey01}.  Concretely,  MTL studies a setting where a set of tasks ( $\mathcal{T} _i$) are made available together upfront.  However,  it cannot handle sequential and dynamic aspects of problems properly. 

In contrast,   online learning is the process of learning sequentially,  however, it does not leverage past experiences like MTL, i.e., it may not consider how past experience can help to enhance the adaptation to a new task.  The Earliest research studies introduced sequential learning  \cite{hannan1957approximation, cesa2006prediction} where tasks are revealed one after another repeatedly. The aim of learning is to learn as independent as possible to attain zero-shot learning  with non task-speciﬁc adaptation.  We argue that neither setting is ideal for studying continual lifelong learning.  MTL deals with learning to learn,  but neglects the sequential and non-stationary aspects of the problem.  Online learning offers an appealing theoretical framework,  but does not generally consider how past experience can accelerate adaptation to a new task.  In this work,  we motivate and present the online MTL problem setting,  where the agent simultaneously uses past experiences in a sequential setting to learn good priors,  and also adapt quickly to the current task at hand.  

The rest of this chapter is organized as follows.  In section 2  emerging challenges in machine learning are discussed .  After that,  applications of MTL using transfer learning are covered.  finally we have promises of MTL.  Figure \ref{fig : General_Structure} represents the overall structure of this study.

\section{Machine learning : challenges and drawbacks}
Prior works on learning process,  regression and optimization problems,  have attempted to learn the behavior of input data,  analyze and categorize it to attain a high performance algorithms.  Machine learning (ML) has been strongly applied to solve supervised and  unsupervised  problems.  ML deploys different algorithms, such as online learning,  multi-task learning and supervised algorithms,  including rule based \cite{weiss1995rule, banda2018advances},  function based \cite{chen2019artificial, iranmehr2019cost},  lazy \cite{agrawal2019integrated},  and bootstrap \cite{poland2019conformal}.   Some of them are used to transform data,  special example would be dimension reduction for optimization,  some to build classiﬁers like supervised algorithms,  others for prediction like regression,  etc.  Machine learning still yields subtle drawbacks for time-varying input data which restrict it to consider properly future and unseen classes to provide general idea and knowledge from data. 

Traditionally, machine learning is a machine learns only input data and predict new data which follow the rule of the equation 
$
  \mathcal{P} _i \times \mathcal{D} \longrightarrow \mathcal{M}$
,  where $ \mathcal{P} _i$ stands for the specific supervised algorithm parameters,  $\mathcal{D}$ represents the space of training data distribution and $\mathcal{M}$ defines the space of generated models which will be applied on test data to evaluate the supervised algorithm performance. 

\begin{figure}[H]
    \centering
    \includegraphics[height=3.6in]{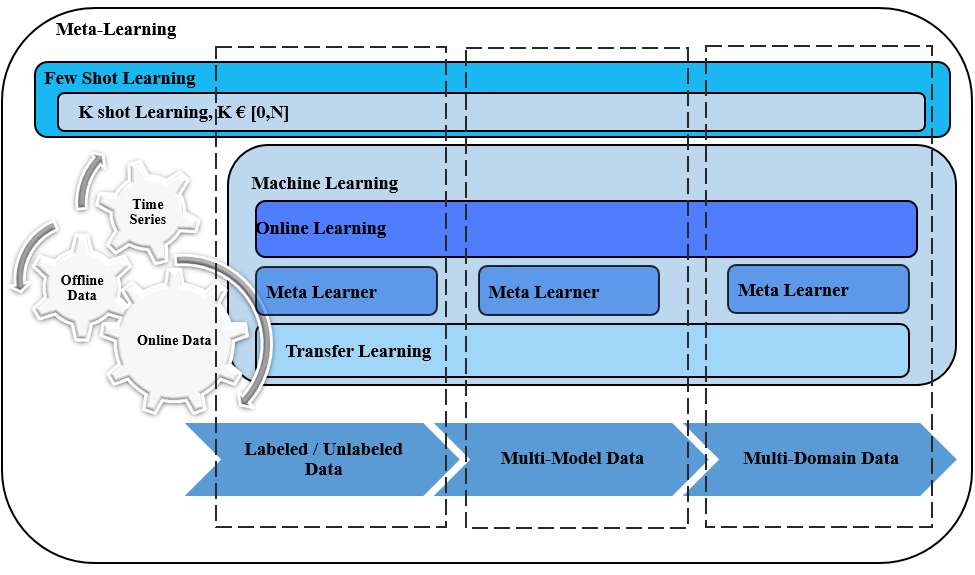}
    \caption{The general overview of Learning against emerging data}
    \label{fig : MTL&ML}
\end{figure}

Figure \ref{fig : MTL&ML} presents few machine learning approaches and algorithms which provide different applications with respect to the wide variety of data such as offline data vs.  online data,  labeled data vs.  unlabeled data,  multi-model data vs.  single model data,  and multi-domain data vs.  single domain data.  As it shows,  machine learning has  critical drawbacks which cannot handle whole data once.  Moreover,  it just considers each data as a new model and each model is separate from previous ones. 

Furthermore,  figure \ref{fig : MTL&ML} depicts the relationship between traditional machine learning and advance machine learning.  In traditional machine learning we have to deal with offline and limited amount of data and the number ground-truth.  However,  in the world of technology,  where data growth have proliferated significantly and are coming from wherever technology exists,  it is very critical to get to know the pattern and rules that govern whole data and learn the trend of the generated data for a specific domain.  For that end,  we need to classify data into three big categories,  time series data,  offline data and online data.  These all categories are shown in three different aspects :  supervised and unsupervised; multi-model data and multi-domain data.  

Machine learning involves transfer learning and online learning,  which is compatible to learn tasks and classes,  which are consequential.  Transfer learning is the theory of transferring knowledge from one task to another and learning from non-randomness.  Meta learner also is one of the bootstrap algorithms which learn data by sampling given data set and generating different data sets.  Meta learner deploys different supervised algorithms and then select a meta learner to give a vote to make a decision about the class of current instance.

\section{Meta-learning algorithms}
\begin{figure}[H]
    \centering
    \includegraphics[height=3. in]{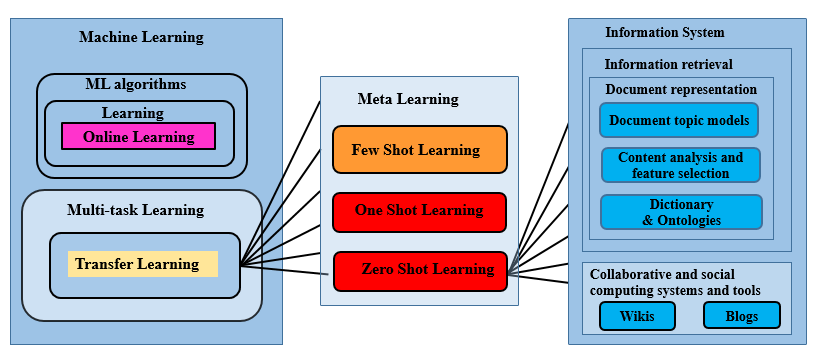}
    \caption{The relation among Machine Learning, Meta Learning,  and Information system
}
    \label{fig : genraloverview}
\end{figure}

Meta-learning (MTL) is firstly presented in  \cite{Schmidhuber1987Learning} and   \cite{bengio1990learning}.  After a decade gap,  lately research studies have tried to deploy  MTL again.   
MTL is a machine which learns the variety of input data.  ML methods need to learn new tasks faster by leveraging previous experiences.  MTL does not consider past experiences separately.
MTL is the process of learning how to learn.  
MTL is an emerging learning algorithms with new challenges and research questions.  It is  an extension of transfer learning,  which is one of the multi-task learning algorithms.  MTL has covers three different aspects as illustrated in Figure \ref{fig : genraloverview}.  Few shot learning (FSL),  one shot learning (OSL) and zero shot learning (ZSL).  FSL and OSL yield highly accurate results as compared with traditional machine learning algorithms.  However,  they still have a critical challenge which limits them from converging to optimal results.  Limits of ZSL have been addressed using domain semantic space,  where includes all information system as presented in Figure \ref{fig : genraloverview}.

\subsection{Model-based MTL}
\begin{figure}[H]
    \centering
    \includegraphics[height=3.4in,  width=6 in]{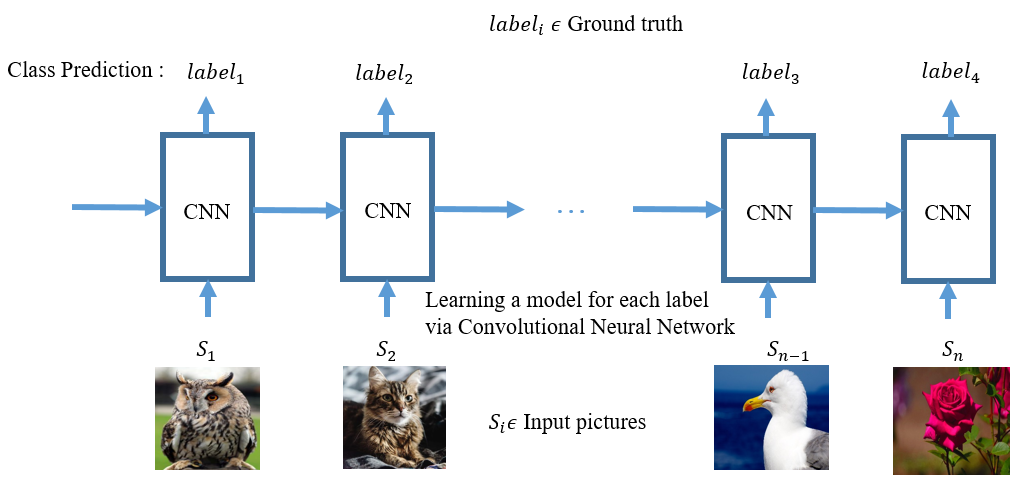}
    \caption{Model-based MTL}
    \label{fig : model-based}
\end{figure}

Model-based MTL depends on a model and no conditional probabilistic method which enable it to be the best match for fast learning model where it updates it's hyper-parameters so fast by training just few examples.  The process of updating their hyper-parameters is done either internal architecture or external meta-learner. 
The concept of model-based MTL is having one neural network interact with sequential neural networks to accelerate the learning process.  In other words,  it tries to learn a model per each label using pixel by pixel value,  according to the figure \ref{fig : model-based}.  In other words,  this model's algorithms try to train a recurrent model like the work presented  \cite{Ravi2017fewshot},  which proposed long short term memory (LSTM).  Hochreiter and Schmidhuber  \cite{hochreiter1997long} proposed for the first time in 1997 the theory of LSTM. 
Model-based algorithms take the data set sequentially and analyze instances one by one.  Since Model-based algorithms leverage RNN for learning,  they become the most least efficient model in comparison with other models.    Nagabandi \emph{et al}  \cite{nagabandi2018deepll} proposed online deep learning using MTL towards Continual adaptation for model-based reinforcement learning.   Santoro \emph{et al}  \cite{santoro2016meta} proposed memory-augmented neural network using MTL. 

\subsection{Metric-based learning}
\begin{figure}[H]
    \centering
    \includegraphics[height=3in,  width=6 in]{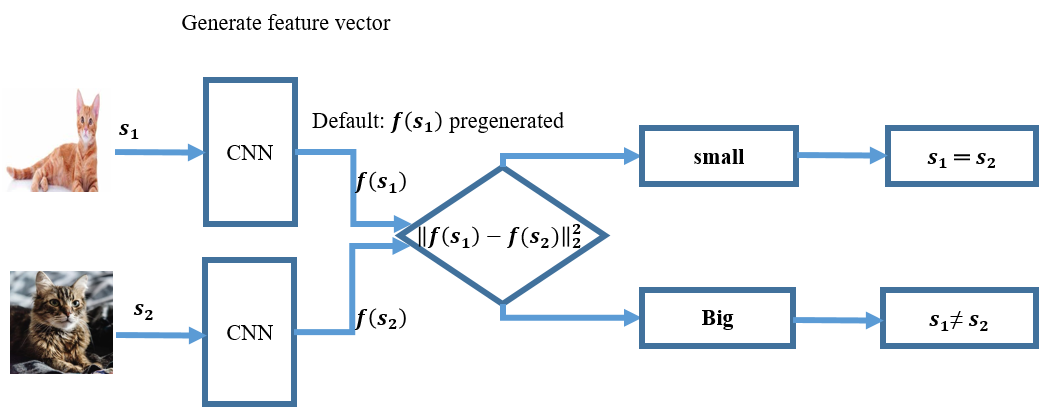}
    \caption{Metric based MTL}
    \label{fig : metric-based}
\end{figure}

Metric-based learning leverages metric space learning,  which leads to efficient data processing and is suitable for few-shot learning.  Lets consider that our goal is image classification.  As model-based learning tries to learn each image pixel by pixel which takes long time and time consuming,  metric-based learning overcomes this limitation by leveraging comparing given two images to the network.  The output per each input yields a vector,  comparing these two vector states that whether they are similar or not.  Figure \ref{fig : metric-based}.  One of the most common application of metric-based learning is  Siamese network presented in  \cite{koch2015siamese}.  Koch \emph{et al} presented Siamese neural network (SNN) for one-shot learning which achieved strong and better results.  The idea behind SNN is that it tries to use twin or half-twin network to compare the input images.  Note that one of the input is already computed and we only need to take the second image and try to go through the layers and compute the vector.  Then,  SNN tries to compute the distance between them,  if the result is small they similar otherwise they are different.  Another application of metric-based learning is  \cite{vinyals2016matching},  where Vinyals \emph{et al} proposed a matching network(MN) for one-shot learning.  

\subsection{Gradient decent-based learning}
This model of MTL is also known as optimization-based model for tuning the parameter ($\theta$).  The idea here is to leverage stochastic gradient-decent (SGD) and for new given sample,  it updates the parametes to be a universal learner.  It may not converg to a local optimal since does not rely on small number of samples. 

Although gradient based learning model works good,  it still has some drawbacks.  Ravi and Larochelle  \cite{Ravi2017fewshot} addressed these problems carefully and  provided LSTM-based MTL to overcome those problems. 
Finn  \cite{finn2017model} presented MAML to improve the accuracy of LSTM-based MTL.  

\begin{table}[H]

    \centering
    \vline
    \begin{tabular}{c|l|}
     \hline
    Abb & Definition\\
     \hline
     CNN & Convolutional neural network \\
     EGNN & Edge-labeling graph neural network\\
     fSL & Few-shot learning \\
     
     GSP & Goal-conditioned skill policy \\
     LSL & Low-shot learning\\
     MAML & Model-agnostic MTL \\
     MANN & Memory-augmented neural network\\
     MIL &Meta Imitation Learning\\
     MTL & Meta-learning \\
     ML & Machine learning \\
     MN & Matching Network\\
     
     OSL & One-shot learning \\
     PN& Prototypical Network\\  
      RN& Relation network\\
       RNN & Recurrent neural network\\
       SAE & Semantic AutoEncoder \\
       SNN & Siamese neural network\\
      TL & Transfer learning \\
         
        ZSL & Zero-shot learning \\
        ZSL-FGVD & ZSL Fine-Grained Visual Descriptions\\
        ZSL-H & Zero-shot learning by mitigating the Hubness problem \\
       ZSL-KT& Zero-shot learning and knowledge transfer\\
      
       \hline
        
    \end{tabular}
  
    \caption{Abbreviation of words}
    \label{tab : AbbreTable}
\end{table}

\begin{landscape}

\begin{table}[H]

    \centering
    \vline
    \begin{tabular}{c|l|l|l|l|l|l|}
     \hline
    Paper & Meta-learning & Proposed method &Meta-learning Models  &Conference / journal &domain & year \\
     \hline
       \cite{finn2017model} & Few-shot learning &   MAML& Gradient decent based &ICML&Image classification& 2017  \\
         \cite{antoniou2018train} & Few-shot learning &   MAML++& Gradient decent based &ICRL&Image classification& 2019  \\
          \cite{finn2018probabilistic} & Few-shot learning & probabilistic MAML & Gradient decent based&NIPS  & Image classification & 2018  \\ 
           \cite{snell2017prototypical} & Few-shot learning & PN &  Metric based  & NIPS& Image classification & 2017   \\
           \cite{Zou2019HML} & Few-shot learning & HML &  Model based  & arXiv& Image classification & 2019   \\
          
            \cite{sung2018learning} & Few-shot learning & RN  &Metric based &  CVPR & Image classification&2018 \\
          
            \cite{Ravi2017fewshot} & Few-shot learning & LSTM-Metalearner  &Gradient decent based&  ICLR & Image classification&2017 \\
              \cite{Kim2019CVPR} & Few-shot learning & EGNN  &Model based &  CVPR & Image classification&2019 \\
              \cite{Wang_2018_CVPR} & Few-shot learning & LSL  &Metric based&  CVPR & Image classification&2019 \\
              \cite{Zou2019HML} & One-shot learning & HML & Model based    & arXiv& Image classification & 2019   \\
          \cite{santoro2016meta}& One-shot learning   & MANN & Model based  & ICML &  Image classification   &2016\\
       \cite{vinyals2016matching}& One-shot learning & MN & Metric based& NIPS & Image classification&2016\\
        \cite{finn2017model} & One-shot learning &   MAML& Gradient decent based &ICML&Image classification& 2017  \\
        \cite{Finn2017oneShot} & One-shot learning& MIL & Gradient decent based&  CoRL & Visual imitation& 2017\\
            \cite{koch2015siamese} & One-shot learning& MIL & Metric based&  ICML & Image recognition& 2015\\
           \cite{reed2016learning} & Zero-shot learning &   ZSL-FGVD  & Metric based& CVPR& Image classification and retrieval&2016\\ 
           \cite{Dinu2015zsl}& Zero-shot learning &  ZSL-H  & Metric based  &  ICLR-workshop & Hubness problem &2015\\
            \cite{sung2018learning} & Zero-shot learning & RN  &Metric based &  CVPR & Image classification&2018 \\   \cite{kodirov2017semantic}& Zero-shot learning& SAE   &Metric based   &  CVPR & Image classification&2017\\
         \cite{sung2018learning}& Zero-shot learning& ZSL-RN   & Metric based&  CVPR & Benchmark classification&2018\\
        
           \cite{choi2019zero} & Zero-shot learning& ZSL-KT &Metric based  &  arXiv & Music classification& 2019\\
            \cite{pathak2018zero} & Zero-shot learning& GSP & Metric based &  CVPR-workshop & Visual imitation & 2018\\
              \cite{pathakICLR18zeroshot} & Zero-shot learning& GSP &Metric based  &  ICLR & Visual imitation& 2018\\

       \hline
        
    \end{tabular}
  
    \caption{An overview of previous studies on Meta-learning. }
    \label{tab : recentStudies}
\end{table}

\end{landscape}

\section{Promises of meta-learning}
\begin{figure}[H]
    \centering
    \includegraphics[height=3.3in]{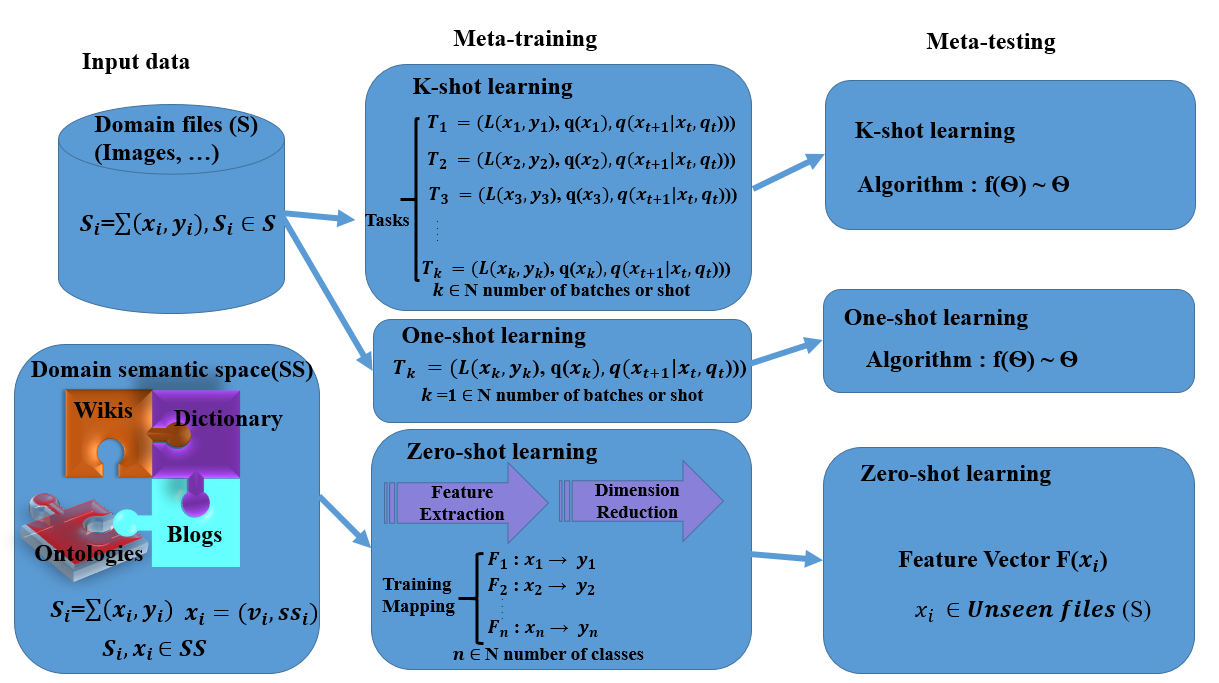}
    \caption{Structure of meta-learning models}
    \label{fig : MTLprocess}
\end{figure}

Learning to learn is an advance process which provides three promises :  one few-shot learning(FSL),  one one-shot learning (OSL),  one zero-shot learning(ZSL).  Figure \ref{fig : MTLprocess} presents a general view of each promises.  we have three layers :  input data,  meta-training,  and meta testing.  Input data for FZL and OSL are the same type,  particularly images for particular image classification aims.  Further,  ZSL becomes an independent learning MTL algorithm which evaluates input data based on domain semantic space and visual information of that domain.

In second layer,  FSL tries to learn k-shot tasks,  which means MTL is training by leveraging k different training data set.  K-shots had generated in advance before learning process have started.  Thus,  MTL is known as a  certain type of bootstrap algorithms,  however,  in k-shot data set we only have specific number of K instances per class.  However,  the bootstrap algorithm tries to split given data set with different rate and would be keeping the ratio of number of classes.  MTL attempts to calculate the loss of each shot using loss function. 
Furthermore,  OSL also ties to learn the task based on k=1 shot learning which means that OSL only has one shot at a moment.  In other words,  when it start bootstraping,  it only select one sample per class as a training set.  Note that OSL represents a special kind of K-shot or few shot learning.  Both FSL and OSL use the equitation below from  \cite{finn2017model}. 
$\mathcal{T}_i=\sum_{i=1}^k(\mathcal{L}(x_i, y_i), q(x_i), q(x_{t+1}|x_t, q_t))$

Further,  in second lay,  ZSL unlike FSL and OSL algorithms ties to work with domain semantic space rather than domain files like images.  The goal here is to find an optimal mapping from semantic space to vector space.  ZSL tries to map given extracted features to a new space called vector space.  

Finally,  last layer stands for the meta-testing which is responsible to predict the given test data and analyze them.  First two algorithms try to predict unseen data using f($\theta$),  however,  zsl attempts to solve the problem by mapping the unseen data to the new vector space. 
\subsection{Few-shot learning}
 The first and one of the most common promises of MTL is few-shot learning (FSL).  Few-shot classification is a specific extension of MTL in supervised learning.  Lake \emph{et al}  \cite{lake2015human} challenged traditional machine learning algorithms by enabling them to learn every concept from one or few shot of that data set.  The idea behind that,  MTL tries to re-sample the given input data set for training using only \textit{K} samples per each class.  In other words,  meta-training process is accomplished by learning \textit{k} shot meta sets which are selected by replacement.  Although few-shot learning outperforms traditional machine learning algorithms,  it has
an explanatory challenge,  called task ambiguity.  This problem happens when a small task,  which is generated from large input data set,  to learn via few-shot learning.  After taking a new task,  which appears too ambiguous to ascertain a single model for that task that covers quite a large number of samples. 

The majority of MTL algorithms leverage few-shot learning.  FSL has decent important extension :  one  Finn \emph{et al} proposed model-agnostic MTL (MAML) \cite{finn2017model},  which adapts to new tasks via gradient descent-based MTL.  
In  \cite{finn2018probabilistic},  Finn \emph{et al} re-sampled models for a new task using a model distribution.  This paper extends MAML to conduct a parameter distribution that is trained through different lower bound.  In  \cite{finn2018probabilistic} Finn et al addressed the ambiguity problem by proposing probabilistic MAML.  

\begin{figure}[H]
    \centering
    \includegraphics[height=3.3in]{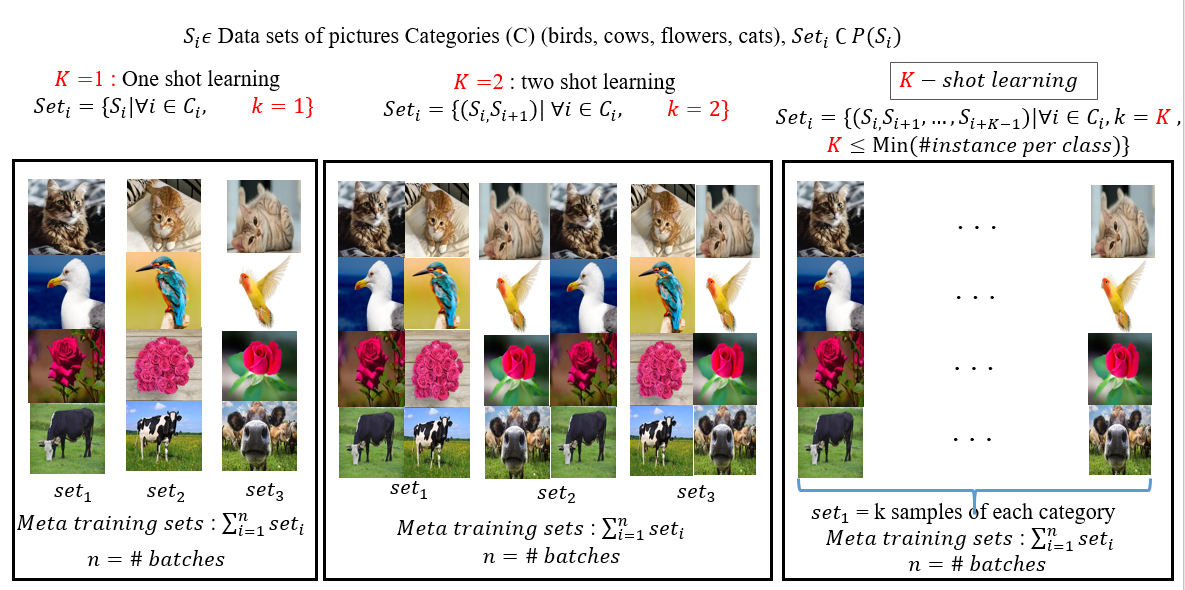}
    \caption{Few-shot learning structure}
    \label{fig : FSL}
\end{figure}

Second important extension is Online learning which is learning process of training data sequentially and continuously.  The next one is online MTL.  Finn \emph{et al}  \cite{finn2019online} proposed online MTL based on the regret-based meta-learner.  Kim \emph{et al}  \cite{Kim2019CVPR} proposed EGNN,  which applied a deep neural network on a certain model,  edge-labeling graph.  Furthermore, 
Sun \emph{et al}  \cite{sun2019meta} proposed an advanced Meta-transfer learning for few-shot learning. 
Zou and Feng  \cite{Zou2019HML} introduced new type of MTL which works based on hierarchy structure,  called Hierarchical MTL(HML).  HML overcomes previous MTL limitation which are limited to the tasks where training sets and identical output structure.  HML enables MTL to optimize adaptability of meta-model to tasks,  that are similar.  Figure\ref{fig : FSL} provides a general view of few shot learning,  one-shot learning and 2-shot learning and generalized k-shot learning. 
\subsection{One shot learning}
One-shot learning (OSL) is a critical challenge in the applications of deep neural networks.  OSL is special type of few-shot learning or k-shot learning in which it choose k=1 shot for training section.  In other words,  when the algorithm starts training,  they only leverage from one instance per class at a time with different batches.  The research studies have been done for one-shot learning are listed as following :  Matching networks  \cite{vinyals2016matching} which is a metric based MTL. 
\begin{figure}[H]
    \centering
    \includegraphics[height=3.3in]{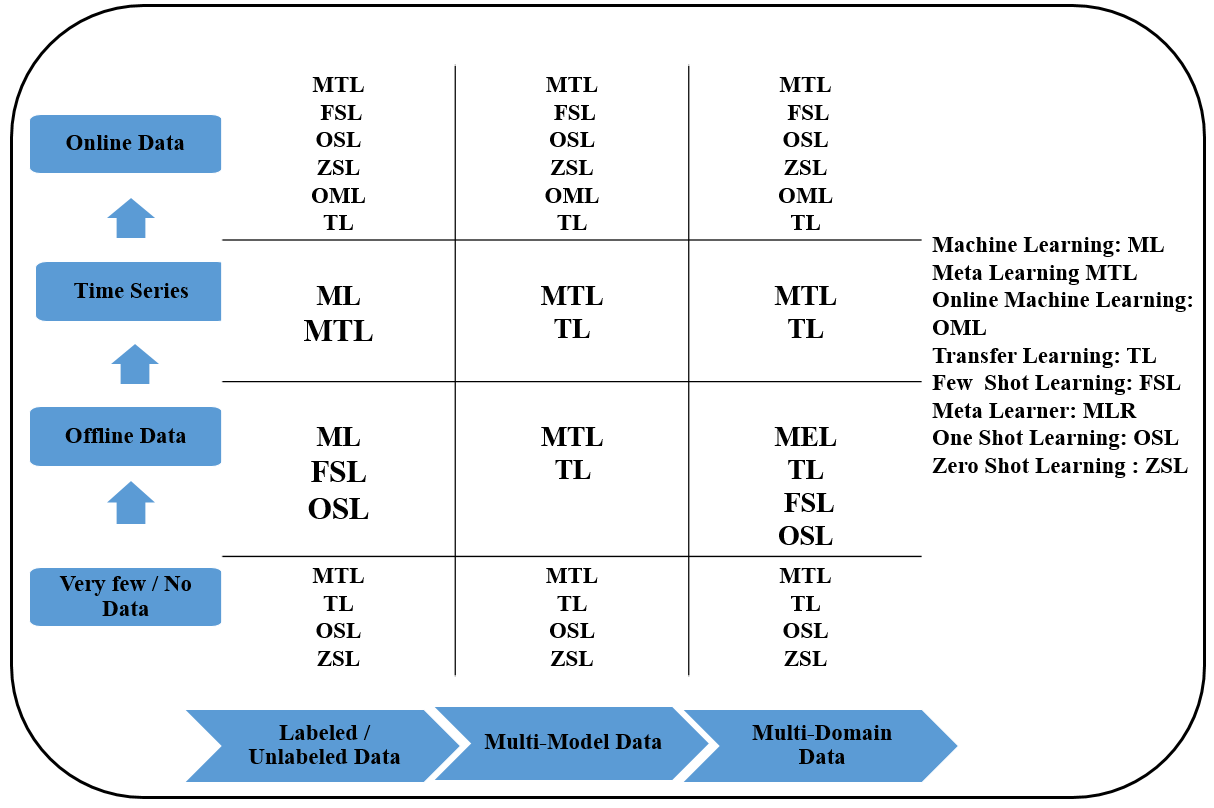}
    \caption{Machine Learning :  ML, 
Meta-Learning:  MTL, 
Online Machine Learning:  OML, 
Transfer Learning:  TL, 
Few  Shot Learning:  FSL, 
Meta Learner:  MLR, 
One Shot Learning:  OSL
}
    \label{fig : Comparison}
\end{figure}
\subsection{zero shot learning}
Zero-shot learning (ZSL) is an emerging paradigm of machine learning which is recently proposed \cite{reed2016learning, sung2018learning, sung2018learning, choi2019zero} to yield a better result than supervised learning algorithms by  covering the their critical limitations,  which work only with a ﬁxed number of classes.  Zero shot learning  is as a joint embedding problem of domain specific and side information,  which includes ontology,  wikis,  dictionary and blogs.  To be certain,  ZSL overcome few-shot learning and one shot learning limitation and promised to yield a result better than FSL and OSL.  The goal is to classify unseen samples of different classes without having a training data set.  This is possible once you have proper information about the domain and classes,  properties and most importantly the functionality of the problem.  The ZSL process is a journey from feature space to a vector space in which it leverages feature extraction and dimension reduction algorithms technically. The feature vector describes shared features among classes.  Reed \emph{et al}  \cite{reed2016learning} applied neural language model to overcome supervised learning limitation. ZSL has been accomplished for visual recognition  \cite{reed2016learning} ,  music classification  \cite{choi2019zero}, and image classification \cite{sung2018learning}. 
More recent methods have been proposed by Kodirov \textit{et al.} \cite{kodirov2017semantic} using auto-encoders for ZSL, Nagabandi \textit{et al.} \cite{nagabandi2018deepll} to deploy MTL for online Learning and by  Finn \textit{et al} \cite{finn2019online} for  online MTL.

\section{Discussion}
Choosing the appropriate type  of data for machine learning algorithms is an important  yet challenging task.  According to   \cite{wang2019atmseer},  it is  crucial to select an optimal algorithm to solve each specific problem with to ensure optimal decision making.  They combined experimental result and interviewed with domain experts.  
It is essential to know where we are,  what are the challenges and what kind of data we have now.  Further,  what is the relationship among emerging data with respect to traditional and modern machine learning algorithms.   Figure \ref{fig : Comparison} presents the information to choose which algorithm are suitable,  compatible and applicable given the specific type of data.

\begin{figure}[H]
    \centering
    \includegraphics[width=7in]{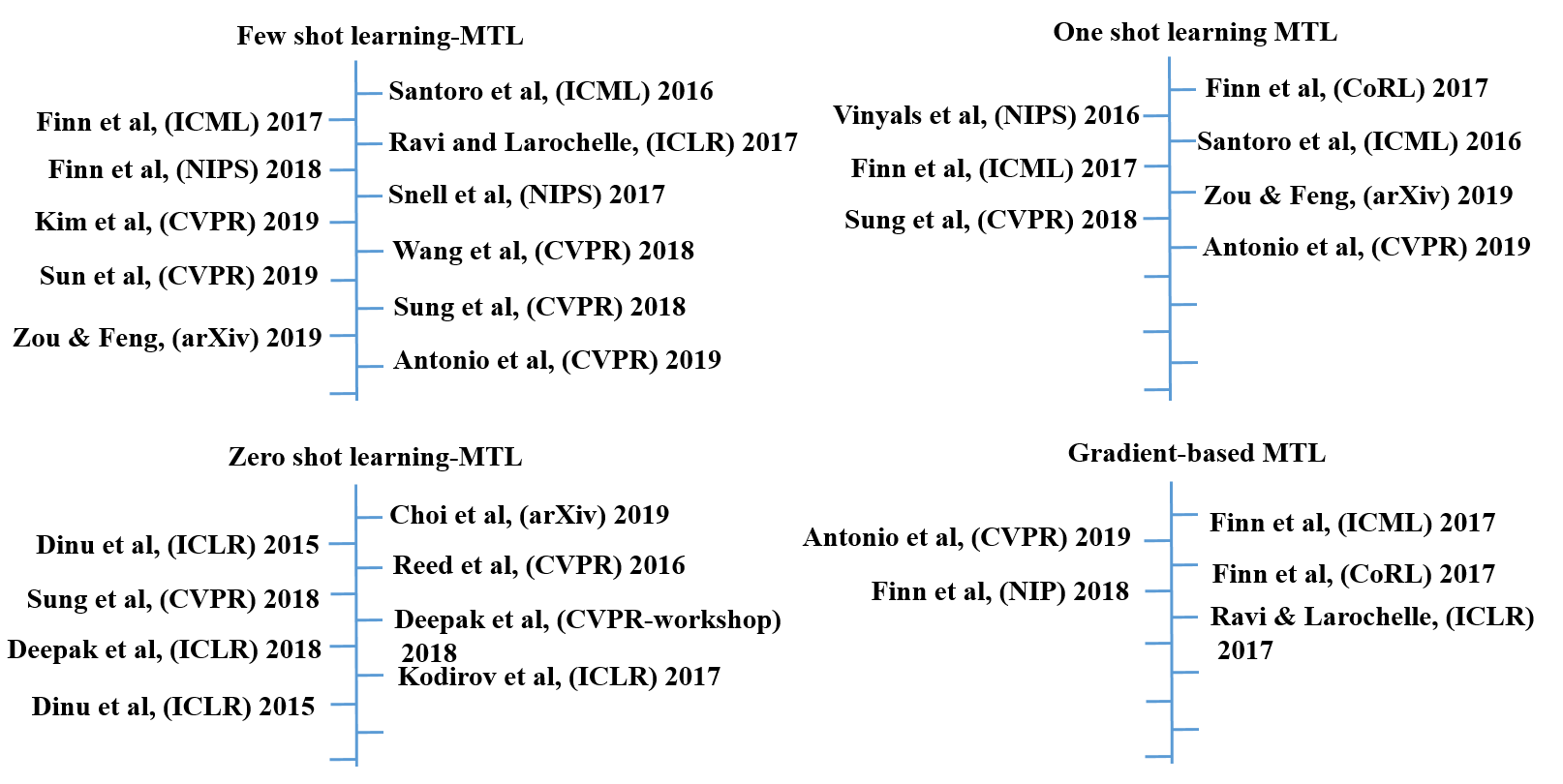}
    \caption{ A brief studies over promises of Meta-learning
}
    \label{fig : review}
\end{figure}

In figure \ref{fig : review}, we have identified some of the publications in top venues. According to our investigation, few-shot learning is one of the most promising areas.  
\section{Conclusion}

Optimizing algorithms to work with offline data is almost ubiquitous in each domain, such as engineering applications.  The majority of studies have determined an optimal way to deal with large-scale problems.  Advancing technologies have people provided data available wherever they have access to internet.  Thus,  it is critical to process continues data which is online and introduce an advance learning algorithm to help scientists to predict future properly.  In this chapter,  we  addressed this problems and investigated an advanced machine learning algorithm to solve them optimally using MTL. 
MTL has three important categories which covers whole research studies have accomplished in this area. One model based, one metric based, one gradient decent based, which also known as optimization method. Further, MTL has three critical extension for emerging data and large-scale problems. The first one, few-shot learning which is practically worked on k-shots of training classes. The second extension is special type of few-shot learning which here we have only one-shot for each training classes. The last one but not the least one is zero-shot learning. Although decent work have been done using FSL and OSL, but ZSL is the promising extension of meta-learning where researchers have no idea about the new classes and no enough data available.

\bibliographystyle{unsrt}
\bibliography{bib.bib}

\begin{thebibliography}{10}

\bibitem{ch1_farid}
F~Ghareh Mohammadi, M~Hadi Amini, and Hamid~R. Arabnia.
\newblock {E}volutionary computation, optimization and learning algorithms for
  data science.
\newblock {\em arXiv preprint arXiv: 1908.08006}, 2019.

\bibitem{ch2_farid}
F~Ghareh Mohammadi, M~Hadi Amini, and Hamid~R. Arabnia.
\newblock Applications of nature-inspired algorithms for dimension {R}eduction:
  Enabling efficient data analytics.
\newblock {\em arXiv preprint arXiv: 1908.1908.08563}, 2019.

\bibitem{Schmidhuber1987Learning}
Jurgen Schmidhuber.
\newblock Evolutionary principles in self-referential learning.
\newblock {\em Diploma thesis}, 1987.

\bibitem{Ravi2017fewshot}
Sachin Ravi and Hugo Larochelle.
\newblock Optimization as a model for few-shot learning.
\newblock In {\em International Conference on Learning Representations (ICLR)},
  pages 281--288, 2017.

\bibitem{melsurvey01}
Ricardo Vilalta and Youssef Drissi.
\newblock A perspective view and survey of meta-learning.
\newblock {\em Artificial intelligence review}, 18(2):77--95, 2002.

\bibitem{hannan1957approximation}
James Hannan.
\newblock Approximation to bayes risk in repeated play.
\newblock {\em Contributions to the Theory of Games}, 3:97--139, 1957.

\bibitem{cesa2006prediction}
Nicolo Cesa-Bianchi and Gabor Lugosi.
\newblock {\em Prediction, learning, and games}.
\newblock Cambridge university press, 2006.

\bibitem{weiss1995rule}
Sholom~M Weiss and Nitin Indurkhya.
\newblock Rule-based machine learning methods for functional prediction.
\newblock {\em Journal of Artificial Intelligence Research}, 3:383--403, 1995.

\bibitem{banda2018advances}
Juan~M Banda, Martin Seneviratne, Tina Hernandez-Boussard, and Nigam~H Shah.
\newblock Advances in electronic phenotyping: from rule-based definitions to
  machine learning models.
\newblock {\em Annual review of biomedical data science}, 1:53--68, 2018.

\bibitem{chen2019artificial}
Mingzhe Chen, Ursula Challita, Walid Saad, Changchuan Yin, and M{\'e}rouane
  Debbah.
\newblock Artificial neural networks-based machine learning for wireless
  networks: A tutorial.
\newblock {\em IEEE Communications Surveys \& Tutorials}, 2019.

\bibitem{iranmehr2019cost}
Arya Iranmehr, Hamed Masnadi-Shirazi, and Nuno Vasconcelos.
\newblock Cost-sensitive support vector machines.
\newblock {\em Neurocomputing}, 343:50--64, 2019.

\bibitem{agrawal2019integrated}
Rashmi Agrawal.
\newblock Integrated parallel k-nearest neighbor algorithm.
\newblock In {\em Smart Intelligent Computing and Applications}, pages
  479--486. Springer, 2019.

\bibitem{poland2019conformal}
David Poland, Slava Rychkov, and Alessandro Vichi.
\newblock The conformal bootstrap: Theory, numerical techniques, and
  applications.
\newblock {\em Reviews of Modern Physics}, 91(1):015002, 2019.

\bibitem{bengio1990learning}
Yoshua Bengio, Samy Bengio, and Jocelyn Cloutier.
\newblock {\em Learning a synaptic learning rule}.
\newblock Universit{\'e} de Montr{\'e}al, D{\'e}partement d'informatique et de
  recherche~…, 1990.

\bibitem{hochreiter1997long}
Sepp Hochreiter and J{\"u}rgen Schmidhuber.
\newblock Long short-term memory.
\newblock {\em Neural computation}, 9(8):1735--1780, 1997.

\bibitem{nagabandi2018deepll}
Anusha Nagabandi, Chelsea Finn, and Sergey Levine.
\newblock Deep online learning via meta-learning: Continual adaptation for
  model-based rl.
\newblock {\em arXiv preprint arXiv:1812.07671}, 2018.

\bibitem{santoro2016meta}
Adam Santoro, Sergey Bartunov, Matthew Botvinick, Daan Wierstra, and Timothy
  Lillicrap.
\newblock Meta-learning with memory-augmented neural networks.
\newblock In {\em International conference on machine learning}, pages
  1842--1850, 2016.

\bibitem{koch2015siamese}
Gregory Koch, Richard Zemel, and Ruslan Salakhutdinov.
\newblock Siamese neural networks for one-shot image recognition.
\newblock In {\em ICML deep learning workshop}, volume~2, 2015.

\bibitem{vinyals2016matching}
Oriol Vinyals, Charles Blundell, Timothy Lillicrap, Daan Wierstra, et~al.
\newblock Matching networks for one shot learning.
\newblock In {\em Advances in neural information processing systems}, pages
  3630--3638, 2016.

\bibitem{finn2017model}
Chelsea Finn, Pieter Abbeel, and Sergey Levine.
\newblock Model-agnostic meta-learning for fast adaptation of deep networks.
\newblock In {\em Proceedings of the 34th International Conference on Machine
  Learning-Volume 70}, pages 1126--1135. JMLR. org, 2017.

\bibitem{antoniou2018train}
Antreas Antoniou, Harrison Edwards, and Amos Storkey.
\newblock How to train your maml.
\newblock {\em arXiv preprint arXiv:1810.09502}, 2018.

\bibitem{finn2018probabilistic}
Chelsea Finn, Kelvin Xu, and Sergey Levine.
\newblock Probabilistic model-agnostic meta-learning.
\newblock In {\em Advances in Neural Information Processing Systems}, pages
  9516--9527, 2018.

\bibitem{snell2017prototypical}
Jake Snell, Kevin Swersky, and Richard Zemel.
\newblock Prototypical networks for few-shot learning.
\newblock In {\em Advances in Neural Information Processing Systems}, pages
  4077--4087, 2017.

\bibitem{Zou2019HML}
Yingtian Zou and Jiashi Feng.
\newblock Hierarchical meta learning.
\newblock {\em arXiv preprint arXiv:1904.09081}, 2019.

\bibitem{sung2018learning}
Flood Sung, Yongxin Yang, Li~Zhang, Tao Xiang, Philip~HS Torr, and Timothy~M
  Hospedales.
\newblock Learning to compare: Relation network for few-shot learning.
\newblock In {\em Proceedings of the IEEE Conference on Computer Vision and
  Pattern Recognition}, pages 1199--1208, 2018.

\bibitem{Kim2019CVPR}
Jongmin Kim, Taesup Kim, Sungwoong Kim, and Chang~D Yoo.
\newblock Edge-labeling graph neural network for few-shot learning.
\newblock In {\em Proceedings of the IEEE Conference on Computer Vision and
  Pattern Recognition}, pages 11--20, 2019.

\bibitem{Wang_2018_CVPR}
Yu-Xiong Wang, Ross Girshick, Martial Hebert, and Bharath Hariharan.
\newblock Low-shot learning from imaginary data.
\newblock In {\em The IEEE Conference on Computer Vision and Pattern
  Recognition (CVPR)}, June 2018.

\bibitem{Finn2017oneShot}
Chelsea Finn, Tianhe Yu, Tianhao Zhang, Pieter Abbeel, and Sergey Levine.
\newblock One-shot visual imitation learning via meta-learning.
\newblock In {\em 1st Conference on Robot Learning (CoRL)}, 2017.

\bibitem{reed2016learning}
Scott Reed, Zeynep Akata, Honglak Lee, and Bernt Schiele.
\newblock Learning deep representations of fine-grained visual descriptions.
\newblock In {\em Proceedings of the IEEE Conference on Computer Vision and
  Pattern Recognition}, pages 49--58, 2016.

\bibitem{Dinu2015zsl}
Angeliki~Lazaridou Georgiana~Dinu and Marco Baroni.
\newblock Improving zero-shot learning by mitigating the hubness problem.
\newblock In {\em Proceedings of the 3rd International Conference on Learning
  Representations (ICLR 2015)}. workshop track, 2015.

\bibitem{kodirov2017semantic}
Elyor Kodirov, Tao Xiang, and Shaogang Gong.
\newblock Semantic autoencoder for zero-shot learning.
\newblock In {\em Proceedings of the IEEE Conference on Computer Vision and
  Pattern Recognition}, pages 3174--3183, 2017.

\bibitem{choi2019zero}
Jeong Choi, Jongpil Lee, Jiyoung Park, and Juhan Nam.
\newblock Zero-shot learning and knowledge transfer in music classification and
  tagging.
\newblock {\em arXiv preprint arXiv:1906.08615}, 2019.

\bibitem{pathak2018zero}
Deepak Pathak, Parsa Mahmoudieh, Guanghao Luo, Pulkit Agrawal, Dian Chen, Yide
  Shentu, Evan Shelhamer, Jitendra Malik, Alexei~A Efros, and Trevor Darrell.
\newblock Zero-shot visual imitation.
\newblock In {\em Proceedings of the IEEE Conference on Computer Vision and
  Pattern Recognition Workshops}, pages 2050--2053, 2018.

\bibitem{pathakICLR18zeroshot}
Deepak Pathak, Parsa Mahmoudieh, Guanghao Luo, Pulkit Agrawal, Dian Chen, Yide
  Shentu, Evan Shelhamer, Jitendra Malik, Alexei~A. Efros, and Trevor Darrell.
\newblock Zero-shot visual imitation.
\newblock In {\em International Conference on Learning Representations (ICLR)},
  2018.

\bibitem{lake2015human}
Brenden~M Lake, Ruslan Salakhutdinov, and Joshua~B Tenenbaum.
\newblock Human-level concept learning through probabilistic program induction.
\newblock {\em Science}, 350(6266):1332--1338, 2015.

\bibitem{finn2019online}
Chelsea Finn, Aravind Rajeswaran, Sham Kakade, and Sergey Levine.
\newblock Online meta-learning.
\newblock {\em arXiv preprint arXiv:1902.08438}, 2019.

\bibitem{sun2019meta}
Qianru Sun, Yaoyao Liu, Tat-Seng Chua, and Bernt Schiele.
\newblock Meta-transfer learning for few-shot learning.
\newblock In {\em Proceedings of the IEEE Conference on Computer Vision and
  Pattern Recognition}, pages 403--412, 2019.

\bibitem{wang2019atmseer}
Qianwen Wang, Yao Ming, Zhihua Jin, Qiaomu Shen, Dongyu Liu, Micah~J Smith,
  Kalyan Veeramachaneni, and Huamin Qu.
\newblock Atmseer: Increasing transparency and controllability in automated
  machine learning.
\newblock In {\em Proceedings of the 2019 CHI Conference on Human Factors in
  Computing Systems}, page 681. ACM, 2019.

\end{thebibliography}
\end{document}